\documentclass[journal]{IEEEtran}
\usepackage{cite}
\usepackage{times}
\usepackage{epsfig}
\usepackage{subfig}
\usepackage{graphicx}
\usepackage{amsmath}
\usepackage{amssymb}
\usepackage{algorithm}               
\usepackage{algorithmic}             
\usepackage{multirow}                

\DeclareMathOperator*{\argmin}{argmin}
\newcommand{\comment}[1]{}

\begin{document}
%
\title{Constructing a Non-Negative Low Rank and Sparse Graph with Data-Adaptive Features}
%
%
%

\author{Liansheng~Zhuang,
        Shenghua~Gao,
        Jinhui~Tang,
        Jingjing~Wang,\\
        Zhouchen~Lin,~\IEEEmembership{Senior~Member},
        and Yi~Ma,~\IEEEmembership{IEEE~Fellow},
\thanks{A preliminary version of this paper was published in CVPR2012~\cite{Zhuang2012CVPR}.
This work is supported by the National Science Foundation of China (No.s 61371192 and 61103134),
and the Science Foundation for Outstanding Young Talent of Anhui Province (BJ2101020001).
Z. Lin is also supported by National Science Foundation of China (No.s 61272341, 61231002 and 61121002).}
\thanks{Liansheng Zhuang, and Jingjing Wang are with University of Science and Technology of China,
Hefei 230027, China. Shenghua Gao and Yi Ma are with ShanghaiTech University, Shanghai, China.
Zhouchen Lin is with Peking University, Beijing, China.
Jinhui Tang is with Nanjing University of Science and Technology, Jiangsu, China.
Liansheng Zhuang is the corresponding author (lszhuang@ustc.edu.cn)} 
}

\maketitle

\begin{abstract}
This paper aims at constructing a good graph for discovering intrinsic data
structures in a semi-supervised learning setting. Firstly, we propose to
build a non-negative low-rank and sparse (referred to as NNLRS)
graph for the given data representation. Specifically, the weights
of edges in the graph are obtained by seeking a nonnegative
low-rank and sparse matrix that represents each data sample as a
linear combination of others. The so-obtained NNLRS-graph can
capture both the global mixture of subspaces structure (by the low
rankness) and the locally linear structure (by the sparseness) of
the data, hence is both generative and discriminative. Secondly,
as good features are extremely important for constructing a good
graph, we propose to learn the data embedding matrix and construct
the graph jointly within one framework, which is termed as NNLRS
with embedded features (referred to as NNLRS-EF). Extensive
experiments on three publicly available datasets demonstrate that
the proposed method outperforms the state-of-the-art graph construction
method by a large margin for both semi-supervised classification and
discriminative analysis, which verifies the effectiveness of our proposed method.

\end{abstract}


\ifCLASSOPTIONpeerreview
\begin{center} \bfseries EDICS Category: 3-BBND \end{center}
\fi
%
\IEEEpeerreviewmaketitle

\section{Introduction}

\IEEEPARstart{I}{n} many big data related applications, e.g.,
image based object recognition, one often lacks sufficient labeled training data
which can be cost and time-prohibitive to obtain. On the other hand, a large
number of unlabeled data are widely available, e.g., from the
Internet. Semi-supervised learning (SSL) can utilize both
labeled samples and richer yet unlabeled samples. Recently, it has
received considerable attention in both computer vision and
machine learning communities~\cite{zhu2006semi}. Among current SSL
methods, graph based SSL is particularly appealing due to its
success in practice and its computational efficiency.

A fundamental problem in graph based SSL is the construction of
a graph from the observed data, which represents the underlying
data structures. Graph based SSL methods treat both
labeled and unlabeled samples from the dataset as nodes in a
graph, and then instantiate edges among these nodes which are
weighted by the affinity between the corresponding pairs of
samples. Label information of the labeled samples can then be
efficiently and effectively propagated to the unlabeled data over
the graph. Most learning methods formalize the propagation process
through a regularized functional on the graph. Despite many forms
used in the literature, the regularizers mainly try to accommodate
the so-called \emph{cluster assumption}~\cite{chen2011TPAMI,
wang2009TPAMI}, i.e., points on the same low-dimensional
smooth structure (such as a cluster, a subspace, or a manifold) are
likely to share the same label. Since one normally does not have (or
care about) an explicit parametric model for the underlying manifolds,
many methods approximate them by constructing
an undirected graph from the observed data points. Therefore,
correctly constructing a graph that can best capture the essential
data structures is critical for all graph based SSL methods
\cite{Belkin2006JMLR,zhu2003icml,zhou2004NIPS,Azran2007icml}.

Lots of efforts have been made to exploit ways of constructing a
good graph for SSL~\cite{john2009IEEE,Daitch2009icml,jebara2009graph,Talukdar2009ecml}.
According to Wright~\cite{john2009IEEE}, an informative graph
should have three characteristics: high discriminating power, low
sparsity and adaptive neighborhood. Guided by these rules, lots of
sparse representation (SR) based graph construction methods have
been proposed \cite{yan2009SDM, cheng2010TIP,He2011CVPR,
Tang2011IST}. However, these SR based methods usually do not
characterize the global structure of data. To overcome this
drawback, Liu \emph{et al.} propose low-rank representation (LRR)
to compute simultaneously the weights in an undirected graph
(referred to as LRR-graph hereafter) that represent the affinities
among all the data samples \cite{liu2010icml,liu2013pami}. But LRR
usually results in a dense graph, and the negative values in LRR
are not physical meaningful for constructing an affinity graph.

On the one hand, by understanding the characteristics of an
informative graph and both the advantages and disadvantages of
previous works, in this paper we propose to harness both sparsity
and low rankness of high-dimensional data to construct an
informative graph. In addition, we will explicitly enforce the
representation to be non-negative so that coefficients of the
representation can be directly converted to graph weights. Such a
graph is called \emph{nonnegative low-rank and sparse graph}
(NNLRS-graph). Specifically, given a set of data points we
represent a data point as a linear combination of other points,
where the coefficients should be both nonnegative and sparse.
Nonnegativity ensures that every data point is in the convex hull
of its neighbors, while sparsity ensures that the involved
neighbors are as few as possible. Moreover, to make data vectors
on the same subspace be clustered into the same cluster, we
enforce the matrix constructed by the coefficient vectors of all
data points to be low-rank. On the other hand, previous works
\cite{john2008TPAMI,gao2013sparse} have shown that by projecting
the data with PCA, the embedded data will greatly facilitate the
subsequent sparse representation and improve the classification
accuracy. Therefore, finding a good data embedding strategy is
also very important for the sparse representation. However, these
previous works do data embedding
\cite{john2008TPAMI,gao2013sparse} and the subsequent sparse
representation separately. Therefore, the learnt features may not
optimize the subsequent representation. Realizing that a good data
representation is important for the good performance of graph
construction and the possible improvement space
\cite{john2008TPAMI,gao2013sparse}, we propose to simultaneously
learn the data embedding matrix and graph, which further improves
the performance of semi-supervised classification.


The contributions of this paper can be summarized as follows:
\begin{itemize}
 \item We propose to learn an NNLRS-graph for
SSL. The sparsity property ensures the
NNLRS-graph to be sparse and capture the local low-dimensional
linear structures of the data. The low-rank characteristic
guarantees that the NNLRS-graph can better capture the global
cluster or subspace structures of the data than SR based
graphs~\cite{He2011CVPR,Tang2011IST}. Thus the robustness of
NNLRS-graph to noise and outliers can be enhanced.
 \item We propose
to simultaneously learn the data embedding and graph. Such a
strategy learns a better data representation, which is more
suitable for building an NNLRS-graph, and consequently enhances
the performance of semi-supervised classification.
\end{itemize}
Extensive experiments demonstrate that the NNLRS-graph
significantly improves the performance of SSL
-- often reducing the error rates by multiple folds!

This article extends its preliminary version~\cite{Zhuang2012CVPR}
in terms of both technique and performance evaluation. First, we
extend our NNLRS framework by jointly learning the data embedding
and NNLRS-graph, which enhances the robustness of NNLRS-graph for
data analysis. Second, we conduct more experiments to evaluate the
proposed algorithms. Third, more details about our methods are
provided and the influence of different parameters is also
discussed in the paper.

The rest of this paper is organized as follows. In
Section~\ref{sec:related_work}, we briefly review works related to
graph construction in SSL. In Section
\ref{sec:nnlr-graph}, we detail the construction of NNLRS-graph,
and in Section \ref{sec:nnlr-feature}, we extend NNLRS by
simultaneously learning the data embedding and NNLRS-graph. We
present experiments and analysis in Section~\ref{sec:exp}.
Finally, we conclude our paper in Section \ref{sec:conclusion}.

\section{Related Work}
\label{sec:related_work}

\textbf{Euclidean distance based methods.} Conceptually, a good
graph should reveal the intrinsic complexity or dimensionality of
data (say through local linear relationship) and also capture
certain global structures of data as a whole (i.e., multiple
clusters, subspaces, or manifolds). Traditional methods (such as
$k$-nearest neighbors and Locally Linear
Reconstruction~\cite{jebara2009graph}) mainly rely on pair-wise
Euclidean distances and construct a graph by a family of
overlapped local patches. The so-obtained graph only captures the
local structures and cannot capture the global structures of the
whole data (i.e. the clusters). Moreover, these methods cannot
produce data-adaptive neighborhoods because of using fixed global
parameters to determinate the graph structure and their weights.
Finally, these methods are sensitive to local data noise and
errors.

\textbf{Sparse representation based methods:} As pointed out in
\cite{john2009IEEE}, sparsity is an important characteristics for
an informative graph. Therefore, lots of researchers propose to
improve the robustness of graph by enforcing sparsity.
Specifically, Yan et al.~\cite{yan2009SDM,cheng2010TIP} proposed
to construct an $\ell_1$-graph via sparse representation
(SR)~\cite{john2008TPAMI} by solving an $\ell_1$ optimization
problem. An $\ell_1$-graph over a data set is derived by encoding
each sample as a sparse representation of the remaining samples,
and automatically selecting the most informative neighbors for
each sample. The neighborhood relationship and graph weights of an
$\ell_1$-graph are simultaneously obtained during the $\ell_1$
optimization in a parameter-free way. Different from traditional
methods, an $\ell_1$-graph explores higher order relationships
among more data points, hence is more powerful and discriminative.
Benefitting from SR, the $\ell_1$-graph is sparse, data-adaptive
and robust to data noise. Following $\ell_1$-graph, other graphs
have also been proposed based on SR in recent
years~\cite{He2011CVPR,Tang2011IST}. However, all these SR based
graphs find the sparsest representation of each sample
\emph{individually}, lacking global constraints on their
solutions. So these methods may be ineffective in capturing the
global structures of data. This drawback may reduce the
performance when the data are grossly corrupted. When not enough
``clean data" are available, SR based methods may not be robust to
noise and outliers~\cite{liu2010icml}.

\comment{ Moreover, since SR is based on the sparsity prior of
high-dimensional data, SR based graphs always tend to capture the local linear relationship of samples.}

\textbf{Low-rank representation based methods.} To capture the
global structure of data, Liu \emph{et al.} propose low-rank
representation (LRR) for data representation and use it to
construct the affinities of an undirected graph (hereafter called
LRR-graph)~\cite{liu2010icml,liu2013pami}. \comment{Different from
SR, LRR uses the low rankness prior of high-dimension data, which
say that: high-dimensional data points often have very low
intrinsic complexity and dimensionality, so that a large
collection of data points may span only an extremely
low-dimensional subspace of the high-dimensional ambient space.}An
LRR-graph jointly obtains the representation of all data under a
global low-rank constraint, thus is better at capturing the global
data structures (such as multiple clusters and subspaces). It has
been proven that, under mild conditions, LRR can correctly
preserve the membership of samples that belong to the same
subspace. However, compared to the $\ell_1$-graph, LRR often
results in a dense graph (see Figure \ref{fig:W}), which is
undesirable for graph-based SSL~\cite{john2009IEEE}. Moreover, as
the coefficients can be negative, LRR allows the data to ``cancel
out each other" by substraction, which lacks physical
interpretation for most visual data. In fact, non-negativity is
more consistent with the biological modeling of visual
data~\cite{Patrik2003Neuro,Lee1999Nature}, and often leads to
better performance for data representation~\cite{Lee1999Nature}
and graph construction~\cite{He2011CVPR}.

\section{Nonnegative Low-Rank and Sparse Graphs}
\label{sec:nnlr-graph}
\subsection{Nonnegative Low-Rank and Sparse Representation}\label{sec:NNLRSR}
Let $X = [x_1,x_2,\cdots,x_n] \in \mathbb{R}^{d \times n}$ be a
matrix whose columns are $n$ data samples drawn from independent
subspaces\footnote{The subspaces $S_1,\cdots,S_k$ are independent
if and only if $\sum_{i=1}^{k}S_i = \bigoplus_{i=1}^{k} S_i$,
where $\bigoplus$ is the direct sum.}. Then each column can be
represented by a linear combination of bases in dictionary $A =
[a_1,a_2,\cdots,a_m]$:
\begin{equation}
\label{eq:nnslrr}
X = AZ,
\end{equation}
where $Z = [z_1,z_2,\cdots,z_n]$ is the coefficient matrix with
each $z_i$ being the \emph{representation} of $x_i$. The
dictionary $A$ is often overcomplete. Hence there can be
infinitely many feasible solutions to problem~(\ref{eq:nnslrr}).
To address this issue, we impose the \emph{most sparsity} and
\emph{lowest rank} criteria, as well as a nonnegative constraint.
That is, we seek a representation $Z$ by solving the following
optimization problem
\begin{equation}
\label{eq:minZ}
\begin{array}{l}
  \min_{Z} \mbox{rank}(Z)+\beta\|Z\|_0, \quad
  \mbox{s.t.} \; \phantom{ {}e} X=AZ, Z \geq 0,
\end{array}
\end{equation}
where $\beta>0$ is a parameter to trade off between low rankness
and sparsity. As observed in~\cite{liu2010icml}, the \emph{low
rankness} criterion is better at capturing the global structure of
data $X$, while the \emph{sparsity} criterion can capture the
local structure of each data vector. The optimal solution
$Z^{\ast}$ is called the nonnegative ``lowest-rank and sparsest''
representation (NNLRSR) of data $X$ with respect to the dictionary
$A$. Each column $z_i^{\ast}$ in $Z^{\ast}$ reveals the
relationship between $x_i$ and the bases in dictionary.

However, solving problem~(\ref{eq:minZ}) is NP-hard. As a common
practice (e.g., \cite{candes2011ACM}) we may solve the following
relaxed convex program instead
\begin{equation}
\label{eq:minZast}
  \begin{array}{l}
    \min_{Z} \|Z\|_\ast + \beta\|Z\|_1, \quad
    \mbox{s.t.} \; \phantom{ {}e} X = AZ, Z \geq 0,
  \end{array}
\end{equation}
where $\|\cdot\|_\ast$ is the nuclear norm of a
matrix~\cite{Cai2010SIAM}, i.e., the sum of the singular values of
the matrix, and $\| \cdot \|_1$ is the $\ell_1$-norm of a matrix,
i.e., the sum of the absolute value of all entries in the matrix.

In real applications, the data are often noisy and even grossly
corrupted. So we have to add a noise term $E$ to
(\ref{eq:nnslrr}). If a fraction of the data vectors are grossly
corrupted, we may reformulate problem~(\ref{eq:minZast}) as
\begin{equation}
\label{equ:optimZ-E}
  \begin{array}{l}
    \min_{Z,E} \|Z\|_\ast+ \beta\|Z\|_1 + \lambda\|E\|_{2,1}, \\
    \mbox{s.t.} \phantom{ {}e} X = AZ + E, Z \geq 0,
  \end{array}
\end{equation}
where \(\|E\|_{2,1}=\sum^n_{j=1}\sqrt{\sum^m_{i=1}([E]_{ij})^2}\)
is called the $\ell_{2,1}$-norm~\cite{Liu2009UAI}, and the
parameter $\lambda>0$ is used to balance the effect of noise,
which is set empirically. The $\ell_{2,1}$-norm encourages the
columns of $E$ to be zero, which assumes that the corruptions are
``sample-specific", i.e., some data vectors are corrupted and the
others are clean. For small Gaussian noise, we can relax the
equality constraint in problem~(\ref{eq:minZ}) as did
in~\cite{candes2010IEEE}. Namely, the Frobenious norm $\|E\|_F$ is
used instead. In this paper, we focus on the $\ell_{2,1}$-norm.

\subsection{LADMAP for Solving NNLRSR}\label{sec:LADMAP}
The NNLRSR problem (\ref{equ:optimZ-E}) could be solved by the
popular alternating direction method
(ADM)~\cite{liu2010icml,Lin2009Report}. However, ADM requires
introducing two auxiliary variables when solving
(\ref{equ:optimZ-E}) and expensive matrix inversions are required
in each iteration. So we adopt a recently developed method called
the linearized alternating direction method with adaptive penalty
(LADMAP)~\cite{Lin2011NIPS} to solve (\ref{equ:optimZ-E}).

We first introduce an auxiliary variable $H$ in order to make the objective function separable:
\begin{equation}
\label{equ:optimZ-W-E}
  \begin{array}{l}
    \min_{Z,H,E} \|Z\|_\ast+ \beta\|H\|_1 + \lambda\|E\|_{2,1}, \\
    \mbox{s.t.} \phantom{ {}e} X = AZ + E, Z = H, H \geq 0.
  \end{array}
\end{equation}
The augmented Lagrangian function of problem~(\ref{equ:optimZ-W-E}) is
\begin{equation}
\label{equ:aLfoptimZ-W-E}
  \begin{array}{rl}
    &L(Z,H,E,Y_1,Y_2,\mu)\\
    =&\|Z\|_\ast+ \beta\|H\|_1 + \lambda\|E\|_{2,1}+\\
    &\left\langle Y_1,X - AZ - E\right\rangle+ \left\langle Y_2,Z - H\right\rangle + \\
    &\frac{\mu}{2}\left(\|X-AZ-E\|_F^2+\|Z - H\|_F^2\right)\\
    =&\|Z\|_\ast+ \beta\|H\|_1 + \lambda\|E\|_{2,1}+\\
    &q(Z,H,E,Y_1,Y_2,\mu)-\frac{1}{2\mu}\left(\|Y_1\|_F^2+\|Y_2\|_F^2\right),
  \end{array}
\end{equation}
where
\begin{equation}
\label{equ:q}
  \begin{array}{rl}
  &q(Z,H,E,Y_1,Y_2,\mu)\\
  =&\frac{\mu}{2}\left(\|X-AZ-E+Y_1/\mu\|_F^2+\|Z-H+Y_2/\mu\|_F^2\right)
  \end{array}
\end{equation}
LADMAP is to update the variables $Z$, $H$ and $E$ alternately, by
minimizing $L$ with other variables fixed, where the quadratic
term $q$ is replaced by its first order approximation at the
previous iterate and a proximal term is then
added~\cite{Lin2011NIPS}. With some algebra, the updating schemes
are as follows.
\begin{equation}
\label{equ:update_ZWE}
  \begin{array}{rl}
  Z_{k+1}=&\argmin\limits_{Z} \|Z\|_*\\
  &+\langle \nabla_Z q(Z_k,H_k,E_k,Y_{1,k},Y_{2,k},\mu_k),Z-Z_k \rangle \\
  &+ \frac{\eta_1\mu_k}{2}\|Z-Z_k\|_F^2\\
  =&\argmin\limits_{Z} \|Z\|_*+ \frac{\eta_1\mu_k}{2}\|Z-Z_k\\
  &+[-A^T(X-AZ_k-E_k+Y_{1,k}/\mu_k)\\
  &+(Z_k-H_k+Y_{2,k}/\mu_k)]/\eta_1\|_F^2\\
  =&\Theta_{(\eta_1\mu_k)^{-1}}(Z_k+[A^T(X-AZ_k-E_k+Y_{1,k}/\mu_k))\\
  &-(Z_k-H_k+Y_{2,k}/\mu_k)]/\eta_1),\\
  H_{k+1}=&\argmin\limits_{H\geq 0} \beta\|H\|_1 + \frac{\mu_k}{2}\|Z_{k+1}-H+Y_{2,k}/\mu_k\|_F^2\\
  =&\max(S_{\beta\mu_k^{-1}}(Z_{k+1}+Y_{2,k}/\mu_k),0),\\
  E_{k+1}=&\argmin\limits_{E} \lambda\|E\|_{2,1}\\
  &+ \frac{\mu_k}{2}\|X-AZ_{k+1}-E+Y_{1,k}/\mu_k\|_F^2\\
  =&\Omega_{\lambda\mu_k^{-1}}(X-AZ_{k+1}+Y_{1,k}/\mu_k),
  \end{array}
\end{equation}
where $\nabla_Z q$ is the partial differential of $q$ with respect to $Z$, $\Theta$, $S$ and $\Omega$ are
the singular value thresholding~\cite{Cai2010SIAM}, shrinkage~\cite{Lin2009Report} and the $l_{2,1}$
minimization operator \cite{liu2010icml}, respectively, and $\eta_1=\|A\|_2^2$. The complete algorithm is
outlined in \textbf{Algorithm}~\ref{alg:nnslrr}.

\renewcommand{\algorithmicrequire}{\textbf{Input:}}
\renewcommand{\algorithmicensure}{\textbf{Output:}}
\begin{algorithm}[tb]
    \caption{Efficient LADMAP Algorithm for NNLRSR}
    \label{alg:nnslrr}
    \begin{algorithmic}[1]
        \REQUIRE
        data matrix $X$, parameters $\beta>0$, $\lambda>0$ \\
        \renewcommand{\algorithmicrequire}{\textbf{Initialize:}}
        \REQUIRE
        $Z_0=H_0=E_0=Y_{1,0}=Y_{2,0}=0$, $\mu_0=0.1$, $\mu_{\max}=10^{10}$, $\rho_0=1.1$, $\varepsilon_1=10^{-6}$, $\varepsilon_2=10^{-2}$, $\eta_1=\|A\|_2^2$, $k=0$.\\
        \WHILE  {$\|X-AZ_k-E_k\|_F/\|X\|_F \geq \varepsilon_1$ or $\mu_k\max(\sqrt{\eta_1}\|Z_{k}-Z_{k-1}\|_F,\|H_{k}-H_{k-1}\|_F,\|E_{k}-E_{k-1}\|_F)/\|X\|_F \geq \varepsilon_2$}

            \STATE Update the variables as (\ref{equ:update_ZWE}).

            \STATE Update Lagrange multipliers as follows:
            $$Y_{1,k+1}=Y_{1,k}+\mu_k(X-AZ_{k+1}-E_{k+1}).$$
            $$Y_{2,k+1}=Y_{2,k}+\mu_k(Z_{k+1}-H_{k+1}).$$
            \vspace{-1em}
            \STATE Update $\mu$ as follows:
            $$\mu_{k+1}=\min(\mu_{\max},\rho\mu_k),\mbox{ where}$$
            \begin{equation*}
            \rho=\left\{
            \begin{array}{ll}
            \rho_0, &\mbox{if } \mu_k\max(\sqrt{\eta_1}\|Z_{k+1}-Z_k\|_F,\\
            &\|H_{k+1}-H_k\|_F,\|E_{k+1}-E_k\|_F)/\|X\|_F\\
            & < \varepsilon_2,\\
            1, &\mbox{otherwise}.
            \end{array}
            \right.
            \end{equation*}
            \STATE Update $k$: $k\leftarrow k+1$.
        \ENDWHILE
        \ENSURE
             an optimal solution $(Z^*, H^*, E^*)$.
    \end{algorithmic}
\end{algorithm}

\subsection{Nonnegative Low Rank and Sparse Graph Construction}
Given a data matrix $X$, we may use the data themselves as the dictionary, i.e., $A$ in
subsections~\ref{sec:NNLRSR} and \ref{sec:LADMAP} is simply chosen as $X$ itself.
With the optimal coefficient matrix $Z^\ast$, we may construct a weighted undirected graph
$G = (V,E)$ associated with a weight matrix $W=\{w_{ij}\}$, where $V = \{v_i\}_{i=1}^n$ is
the vertex set, each node $v_i$ corresponding to a data point $x_i$, and $E = \{e_{ij}\}$ is
the edge set, each edge $e_{ij}$ associating nodes $v_i$ and $v_j$ with a weight $w_{ij}$.
As the vertex set $V$ is given, the problem of graph construction is to determine the graph
weight matrix $W$.

Since each data point is represented by other samples, a column
$z_i^{\ast}$ of $Z^\ast$ naturally characterizes how other samples
contribute to the reconstruction of $x_i$. Such information is
useful for recovering the clustering relation among samples. The
sparse constraint ensures that each sample is associated with only
a few samples, so that the graph derived from $Z^{\ast}$ is
naturally sparse. The low rank constraint guarantees that the
coefficients of samples coming from the same subspace are highly
correlated and fall into the same cluster, so that $Z^{\ast}$ can
capture the global structure (i.e. clusters) of the whole data.
Note here that, since each sample can be used to represent itself,
there always exist feasible solutions even when the data sampling
is insufficient, which is different from SR.

After obtaining $Z^{\ast}$, we can derive the graph adjacency
structure and graph weight matrix from it. In practice, due to
data noise the coefficient vector $z_i^{\ast}$ of point $x_i$ is
often dense with small values. As we are only interested in the
global structure of the data, we can normalize the reconstruction
coefficients of each sample (i.e. $z_i^{\ast} =
z_i^{\ast}/\|z_i^{\ast}\|_2$) and make those coefficients under a
given threshold zeros. After that, we can obtain a sparse
$\hat{Z}^{\ast}$ and define the graph weight matrix $W$ as
\begin{equation}
\label{eq:W}
W = (\hat{Z}^{\ast} + ({\hat{Z}^{\ast}})^T) /2.
\end{equation}
The method for constructing an NNLRS-graph is summarized in
\textbf{Algorithm}~\ref{alg:nnlrgraph}.
\renewcommand{\algorithmicrequire}{\textbf{Input:}}
\renewcommand{\algorithmicensure}{\textbf{Output:}}
\begin{algorithm}[t]
\caption{Nonnegative low rank and sparse graph construction}
\label{alg:nnlrgraph}
\begin{algorithmic}[1]
   \REQUIRE
    Data matrix $X = [x_1,x_2,\cdots,x_n] \in \mathbb{R}^{d \times n}$, \\
    \quad regularization parameters $\beta>0$ and $\lambda>0$, threshold
    $\theta>0$.

   \renewcommand{\algorithmicrequire}{\textbf{Steps:}}
   \REQUIRE

   \STATE Normalize all the samples \( \hat{x}_i = x_i/\|x_i\|_2\) to obtain $\hat{X} = \{\hat{x}_1, \hat{x}_2,\cdots,\hat{x}_n\}$. \\

   \STATE Solve the following problem using Algorithm~\ref{alg:nnslrr},
    \[
       \begin{array}{l}
        \min_{Z,E} \ {\|Z\|_{\ast} + \beta \|Z\|_1 + \lambda \|E\|_{2,1}} \\
        \mbox{s.t.} \ \hat{X} = \hat{X}Z + E, Z \geq 0
      \end{array}
    \]
    and obtain an optimal solution $(Z^{\ast}, E^{\ast})$. \\

    \STATE Normalize all column vectors of $Z^{\ast}$ by $z_i^{\ast} = z_i^{\ast}/\|z_i^{\ast}\|_2$,
    make small values under the given threshold $\theta$ zeros by
    \[
    \hat{z}_{ij}^{\ast} = \left\{
        \begin{array}{ll}
            z_{ij}^{\ast}, & \mbox{if}\ z_{ij}^{\ast} \geq \theta, \\
            0, & \mbox{otherwise},
        \end{array}
    \right.
    \]
   and obtain a sparse $\hat{Z}^{\ast}$.

   \STATE Construct the graph weight matrix $W$ by
   \[
       W = ( \hat{Z}^{\ast} + (\hat{Z}^{\ast})^T) / 2.
   \]

   \ENSURE  The weight matrix $W$ of NNLRS-graph.
\end{algorithmic}
\end{algorithm}

\section{Jointly learning data representation and NNLRS-graph}
\label{sec:nnlr-feature}

The quality of data representation will greatly affect the quality
of graph. The data representation which is robust to the data
variance improves the robustness of the graph, and subsequently
improves the performance of SSL. To improve
the data representation, lots of endeavors have been
made~\cite{jolliffe2005principal,he2003locality}. For face data,
the commonly used data representation are EigenFaces
\cite{turk1991eigenfaces}, LaplacianFaces \cite{zhang2005face},
FisherFaces \cite{belhumeur1997eigenfaces}, and RandomFaces
\cite{john2008TPAMI}. As shown in \cite{john2008TPAMI}, these
representation strategies greatly improve the data representation
quality, and improve the classification accuracy. However, the
data embedding and the subsequent sparse representation are
conducted separately in \cite{john2008TPAMI}, and a data embedding
method in the previous step may not be the most suitable for the
subsequent sparse representation.

Other than doing the data embedding and learning the NNLRS graph
separately, we propose to learn the data representation and graph
simultaneously to make the learnt data representation more
suitable for the construction of NNLRS-graph. We first denote the
data projection matrix as $P$. Similar to~\cite{le2011ica}, we
want the projected data to preserve the data information as much
as possible. So we aim at minimizing $\|X-P^TPX\|_F^2$. By
plugging the learning of $P$ into the NNLRS-graph construction
framework, we arrive at the following formulation:
\begin{equation}
\label{equ:NNLRS_EF}
  \begin{array}{l}
    \min_{Z,E,P} \|Z\|_\ast + \beta\|Z\|_1 + \lambda\|E\|_{2,1} + \gamma\|X-P^TPX\|_F^2, \\
    \mbox{s.t.} \phantom{ {}e} PX = PXZ + E, Z \geq 0,
  \end{array}
\end{equation}
where $\gamma$ is a parameter to balance the reconstruction error,
which is set empirically. For simplification, we term this
formulation as NNLRS with embedded feature (referred to as
NNLRS-EF). However, the objective function of NNLRS-EF is not
convex. Therefore it is inappropriate to optimize all the
variables in problem (\ref{equ:NNLRS_EF}) simultaneously.
Following the commonly used strategy in dictionary
learning~\cite{lee2007efficient,elad2006image}, we alternatively
update the unknown variables. Specifically, we first optimize the
above objective w.r.t. $Z$ and $E$ by fixing $P$, then we update
$P$ and $E$ while fixing $Z$.

When $P$ is fixed, \eqref{equ:NNLRS_EF} reduces to
\begin{equation}
\label{equ:EZ}
  \begin{array}{l}
    \min_{Z,E} \|Z\|_\ast + \beta\|Z\|_1 + \lambda\|E\|_{2,1}  \\
    \mbox{s.t.} \phantom{ {}e} PX = PXZ + E, Z \geq 0,
  \end{array}
\end{equation}
We use LADMAP to solve for $Z$ and $E$. By introducing an
auxiliary variable $H$, we obtain the following augmented
Lagrangian function:
\begin{equation}
\label{equ:aLfoptimZ-W-E'}
  \begin{array}{rl}
    &\tilde{L}(Z,H,E,Y_1,Y_2,\mu)\\
    =&\|Z\|_\ast+ \beta\|H\|_1 + \lambda\|E\|_{2,1}  + \\
    &\left\langle Y_1,PX - PXZ - E\right\rangle+ \left\langle Y_2,Z - H\right\rangle + \\
    &\frac{\mu}{2}\left(\|PX-PXZ-E\|_F^2+\|Z-H\|_F^2\right)\\
    =&\|Z\|_\ast+ \beta\|H\|_1 + \lambda\|E\|_{2,1} +  \\
     & \tilde{q_1}(Z,H,E,Y_1,Y_2,\mu)-\frac{1}{2\mu}\left(\|Y_1\|_F^2+\|Y_2\|_F^2\right),
  \end{array}
\end{equation}
where
\begin{equation}
\label{equ:q1}
  \begin{array}{rl}
  &\tilde{q_1}(Z,H,E,Y_1,Y_2,\mu)\\
  =&\frac{\mu}{2}\left(\|PX-PXZ-E+Y_1/\mu\|_F^2+\|Z-H+Y_2/\mu\|_F^2\right)
  \end{array}
\end{equation}
Then we can apply Algorithm~\ref{alg:nnslrr} to (\ref{equ:EZ}) by
simply replacing $X$ and $A$ in Algorithm~\ref{alg:nnslrr} with
$PX$.

After updating $Z$ and $E$, we only fix $Z$. So
\eqref{equ:NNLRS_EF} reduces to the following problem
\begin{equation}
\label{equ:EP}
  \begin{array}{l}
    \min_{E,P}\lambda\|E\|_{2,1} + \gamma\|X-P^TPX\|_F^2, \\
    \mbox{s.t.} \phantom{ {}e} PX = PXZ + E, Z \geq 0.
  \end{array}
\end{equation}
We can solve the above problem with inexact
ALM~\cite{Lin2009Report}. The augmented Lagrange function is
\begin{equation}
\label{equ:aLfoptimPE}
  \begin{array}{rl}
    &\tilde{L}(P,E,\mu)\\
    =& \lambda\|E\|_{2,1} + \gamma\|X-P^TPX\|_F^2 \\
    &+\frac{\mu}{2} \|PX-PXZ-E\|_F^2 + \left\langle Y_1,PX-PXZ-E\right\rangle.\\
  \end{array}
\end{equation}
By minimizing $\tilde{L}(P,E,\mu)$ with other variables fixed, we
can update the variables $E$ and $P$ alternately as
follows\footnote{we solve the subproblem for $P$ with the
L-BFGS~\cite{AYNg2011nips} algorithm. The codes can be found at
http://www.di.ens.fr/~mschmidt/Software/minFunc.html}.
\begin{equation}
\label{equ:update_EP}
  \begin{array}{rl}
  E_{k+1}=&\argmin\limits_{E} \lambda\|E\|_{2,1}\\
  &+ \frac{\mu_k}{2}\|P_kX-P_kXZ-E+Y_{1,k}/\mu_k\|_F^2\\
  =&\Omega_{\lambda\mu_k^{-1}}(P_kX-P_kXZ+Y_{1,k}/\mu_k), \\
  P_{k+1}=&\argmin\limits_{P} \gamma\|X-P^TPX\|_F^2 \\
  &+ \frac{\mu_k}{2}\|PX-PXZ-E_{k+1}+Y_{1,k}/{\mu_k}\|_F^2.\\
  \end{array}
\end{equation}
\renewcommand{\algorithmicrequire}{\textbf{Input:}}
\renewcommand{\algorithmicensure}{\textbf{Output:}}
\begin{algorithm}[tb]
    \caption{Efficient Inexact ALM Algorithm for problem~(\ref{equ:EP})}
    \label{alg:ep}
    \begin{algorithmic}[1]
        \REQUIRE
        data matrix $X$, parameters $\lambda>0$, $\gamma>0$, $Z$ \\
        \renewcommand{\algorithmicrequire}{\textbf{Initialize:}}
        \REQUIRE
        $E_0=Y_{1,0}=0$, $P_0 = I$, $\mu_0=0.1$, $\mu_{\max}=10^{10}$, $\rho=1.1$, $\varepsilon_1=10^{-6}$, $\varepsilon_2=10^{-3}$, $k=0$.\\
        \WHILE  {$\|P_kX-P_kXZ-E_k\|_F/\|P_kX\|_F \geq \varepsilon_1$ or $\|E_{k}-E_{k-1}\|_F/\|P_kX\|_F\geq \varepsilon_2$ or $\|P_{k}-P_{k-1}\|_F/\|P_kX\|_F\geq \varepsilon_2$}
            \STATE Update the variables as (\ref{equ:update_EP}).
            \STATE Update the Lagrange multiplier as follows:
            $$Y_{1,k+1}=Y_{1,k}+\mu_k(P_{k+1}X-P_{k+1}XZ-E_{k+1}).$$
            \STATE Update $\mu$ as follows:
            $$\mu_{k+1}=\min(\mu_{\max},\rho\mu_k)$$
            \STATE Update $k$: $k\leftarrow k+1$.
        \ENDWHILE
        \ENSURE
             an optimal solution $(E^*, P^*)$.
    \end{algorithmic}
\end{algorithm}

We alternatively solve problem~(\ref{equ:EZ}) and
problem~(\ref{equ:EP}) until convergence. The whole process of the
optimization of NNLRS-EF is summarized in Algorithm
\ref{alg:nnlrs-ef}. After getting the optimal solution $Z^*$, we
use the same strategy as that of NNLRS-graph to construct a graph.

\renewcommand{\algorithmicrequire}{\textbf{Input:}}
\renewcommand{\algorithmicensure}{\textbf{Output:}}
\begin{algorithm}[tb]
    \caption{Optimization Algorithm for NNLRS-EF \eqref{equ:NNLRS_EF}}
    \label{alg:nnlrs-ef}
    \begin{algorithmic}[1]
        \REQUIRE
        data matrix $X$, parameters $\beta>0$, $\lambda>0$, $\gamma>0$, $\varepsilon_3>0$\\
        \WHILE  {difference between successive $Z$, $P$ or $E$ is greater than $\varepsilon_3$}
            \STATE Update the variables $E$, and $Z$ by solving problem~(\ref{equ:EZ}) with \textbf{Algorithm~\ref{alg:nnslrr}}.
            \STATE Update $E$ and $P$ by solving problem~(\ref{equ:EP}) with \textbf{Algorithm~\ref{alg:ep}}
        \ENDWHILE
        \ENSURE
             an optimal solution $(Z^*, P^*, E^*)$.
    \end{algorithmic}
\end{algorithm}

\section{Experiments}
\label{sec:exp} In this section, we evaluate the performance of
our proposed methods on publicly available databases, and compare
them with currently popular graphs under the same SSL setting. Two
typical SSL tasks are considered,
semi-supervised classification and semi-supervised dimensionality
reduction. All algorithms are implemented with Matlab 2010. All
experiments are run 50 times (unless otherwise stated) on a server
with an Intel Xeon5680 8-Core 3.50GHz processor and 16GB memory.

\subsection{Experiment Setup}
\textbf{Databases:} We test our proposed methods on three public
databases\footnote{Available at
http://www.zjucadcg.cn/dengcai/Data/} for evaluation: YaleB, PIE,
and USPS. YaleB and PIE are face databases and USPS is a
hand-written digit database. We choose them because NNLRS-graph
aims at extracting a linear subspace structure of data. So we have
to select databases that roughly have linear subspace structures.
It is worth pointing out that these datasets are commonly used in
the SSL literature. Existing methods have
achieved rather decent results on these data sets. So surpassing
them on these three data sets is very challenging and convincing
enough to justify the advantages of our method.

\begin{itemize}
\item \textbf{The YaleB Database}: This face database has 38
individuals, each subject having about 64 near frontal images
under different illuminations. We simply use the cropped images of
first 15 individuals, and resize them to $32 \times 32$ pixels.

\item \textbf{The PIE Database}: This face database contains 41368
images of 68 subjects with different poses, illumination and
expressions. We select the first 15 subjects and only use their
images in five near frontal poses (C05, C07, C09, C27, C29) and
under different illuminations and expressions. Each image is
manually cropped and normalized to a size of $32 \times 32$
pixels.

\item \textbf{The USPS Database}: This handwritten digit database
contains 9298 handwritten digit images in total, each having $16
\times 16$ pixels. We only use the images of digits 1, 2, 3 and 4
as four classes, each having 1269, 926, 824 and 852 samples,
respectively. So there are 3874 images in total.
\end{itemize}

\begin{figure}[tb]
\centering
\DeclareGraphicsExtensions{.eps,.mps,.pdf,.jpg,.png}
\includegraphics[width = 0.45\textwidth]{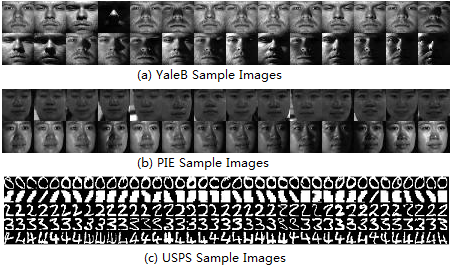}
\caption{Sample images used in our experiments.}
\label{fig:sample}
\end{figure}

Fig.~\ref{fig:sample} shows the sample images of the three
databases. As suggested by~\cite{john2008TPAMI}, we normalize the
samples so that they have a unit $\ell_2$ norm.

\textbf{Comparison Methods:} We compare our proposed graph
construction methods with the following baseline methods:
\begin{itemize}
\item \textbf{$k$NN-graph:} We adopt Euclidean distance as the
similarity measure, and use a Gaussian kernel to re-weight the
edges. The Gaussian kernel parameter $\sigma$ is set to 1. There
are two configurations for constructing graphs, denoted as
\textbf{$k$NN0} and \textbf{$k$NN1}, where the numbers of nearest
neighbors are set to 5 and 8, respectively.

\item \textbf{LLE-graph~\cite{wang2009TPAMI}:} Following the lines
of~\cite{wang2009TPAMI}, we construct two LLE-graphs, denoted as
\textbf{LLE0} and \textbf{LLE1}, where the numbers of nearest
neighbors are 8 and 10, respectively. Since the weights $W$ of
LLE-graph may be negative and asymmetric, similar
to~\cite{cheng2010TIP} we symmetrize them by $W = (|W| +
|W^T|)/2$.

\item \textbf{$\ell_1$-graph~\cite{cheng2010TIP}:} Following the
lines of~\cite{cheng2010TIP}, we construct the $\ell_1$-graph.
Since the graph weights $W$ of $\ell_1$-graph is asymmetric, we
also symmetrize it as suggested in~\cite{cheng2010TIP}.

\item \textbf{SPG~\cite{He2011CVPR}:} In essence, the SPG problem
is a lasso problem with the nonnegativity constraint, without
considering corruption errors. Here we use an existing
toolbox\footnote{http://sparselab.stanford.edu/} to solve the
lasso problem, and construct the SPG graph following the lines
of~\cite{He2011CVPR}.

\item \textbf{LRR-graph:} Following~\cite{liu2010icml}, we
construct the LLR-graph, and symmetrize it as we do for
$\ell_1$-graph. The parameters of LRR are the same as those
in~\cite{liu2010icml}.

\item \textbf{NNLRS-graph:} Similar to our graph construction
methods, for our NNLRS-graph and its extension, we empirically
tune the regularization parameters according to different data
sets, so as to achieve the best performance. Without loss of
generality, we fix the reduced dimensionality to 100 in all our
experiments.
\end{itemize}

\subsection{Semi-supervised Classification}

In this subsection, we carry out the classification experiments on
the above databases using the existing graph based SSL frameworks.
We select two popular methods, \emph{Gaussian Harmonic Function}
(GHF)~\cite{zhu2003icml} and \emph{Local and Global Consistency}
(LGC)~\cite{zhou2004NIPS} to compare the effectiveness of
different graphs. Let $Y = [Y_l \ Y_u]^T \in \mathbb{R}^{|V|
\times c}$ be a label matrix, where $Y_{ij} =1$ if sample $x_i$ is
associated with label $j$ for $j \in \{1,2,\cdots,c\}$ and
$Y_{ij}=0$ otherwise. Both GHF and LGC realize the label
propagation by learning a classification function $F = [F_l \
F_u]^T \in \mathbb{R}^{|V| \times c}$. They utilize the graph and
the known labels to recover the continuous classification function
by optimizing different predefined energy functions. GHF combines
Gaussian random fields and harmonic function for optimizing the
following cost on a weighted graph to recover the classification
function $F$:
\begin{equation}
\label{eq:GHF}
\min_{F \in \mathbb{R}^{|V| \times c}} tr(F^T L_W F), \; \mbox{s.t.} \; L_W F_u=0, F_l=Y_l,
\end{equation}
where $L_W=D-W$ is the graph Laplacian, in which $D$ is a diagonal matrix with $D_{ii} = \sum_j W_{ij}$.
Instead of clamping the classification function on labeled nodes by setting hard constraints $F_l = Y_l$,
LGC introduces an elastic fitness term as follows:
\begin{equation}
\label{eq:LGC}
\min_{F \in \mathbb{R}^{|V| \times c}} tr\{F^T \tilde{L}_W F + \mu (F - Y)^T(F - Y)\},
\end{equation}
where $\mu \in [0, +\infty)$ trades off between the local fitting
and the global smoothness of the function $F$, and $\tilde{L}_W$
is the normalized graph Laplacian $\tilde{L}_W = D^{-1/2} L_W
D^{-1/2}$. In our experiments, we simply fix $\mu = 0.99$.

\begin{figure}[tb]
\centering
\subfloat[LRR-graph Weights]{
\label{fig:W_lrr}
\begin{minipage}[t]{0.23\textwidth}
   \centering
   \includegraphics[width = 1\textwidth]{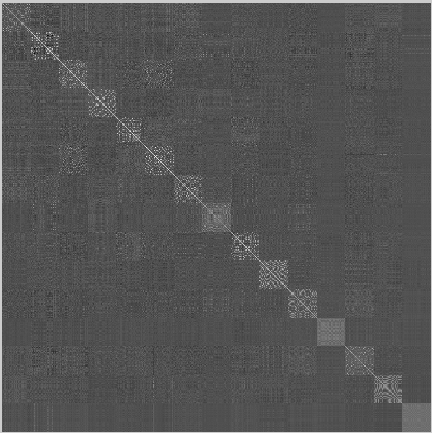}
   \vspace{-3mm}
\end{minipage}
}
\subfloat[$\ell_1$-graph Weights]{
\label{fig:W_L1}
\begin{minipage}[t]{0.23\textwidth}
   \centering
   \includegraphics[width = 1\textwidth]{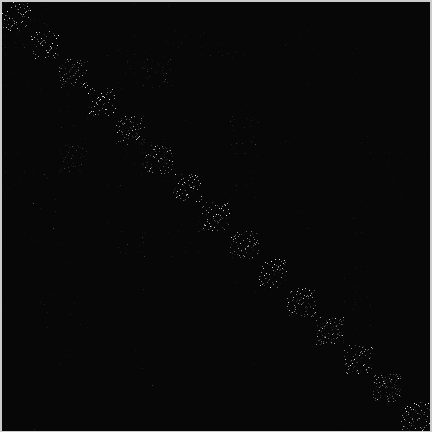}
    \vspace{-3mm}
\end{minipage}
}
\caption{Visualization of different graph weights $W$ on the YaleB
face database.}
\label{fig:W}
\end{figure}

We combine different graphs with these two SSL frameworks, and
quantitatively evaluate their performance by following the
approaches in~\cite{yan2009SDM,cheng2010TIP,john2009IEEE,
He2011CVPR}. For the YaleB and PIE databases, we randomly select
50 images from each subject as our data sets in each run. Among
these 50 images, images are randomly labeled. For the USPS
database, we randomly select 200 images for each category, and
randomly label them. Different from~\cite{yan2009SDM,He2011CVPR},
the percentage of labeled samples ranges from 10\% to 60\%,
instead of ranging from 50\% to 80\%. This is because the goal of
SSL is to reduce the number of labeled images. So we are more
interested in the performance of SSL methods with low labeling
percentages. The final results are reported in
Tables~\ref{tab:GHF} and~\ref{tab:LGC}, respectively. From these
results, we can observe that:

\begin{enumerate}
 \item [1)] In most cases, NNLRS-graph and its extension (i.e.
NNLRS-EF) consistently achieve the lowest classification error
rates compared to the other graphs, even at low labeling
percentages. In many cases, the improvements are rather
significant -- cutting the error rates by multiple folds! This
suggests that NNLRS-graph and its extension are more informative
and thus more suitable for semi-supervised classification.
 \item [2)] Compared with NNLRS-graph, NNLRS-EF also has a significant
improvements in most cases. This demonstrates that a good data
representation can markedly improve the performance of graph
construction methods. This is because good representation is
robust to data noise and helps to reveal the relationship among
data points.
 \item [3)] Though LRR always results in dense graphs,
the performance of LRR-graph based SSL methods is not always
inferior to that of $\ell_1$-graph based SSL methods. On the
contrary, LRR-graph performs as well as $\ell_1$-graph in many
cases. As illustrated in Fig.~\ref{fig:W}, the weights $W$ of
LRR-graph on the YaleB database is denser than that of
$\ell_1$-graph. However, LRR-graph outperforms $\ell_1$-graph in
all cases. This proves that the low rankness property of
high-dimensional data is as important as the sparsity property for
graph construction.
\end{enumerate}

\subsection{Semi-supervised Discriminant Analysis}
To further examine the effectiveness of NNLRS-graph, we use
NNLRS-graph for semi-supervised dimensionality reduction (SSDR),
and take semi-supervised discriminant analysis
(SDA)~\cite{Cai2007ICCV} for instance. We use SDA to do face
recognition on the face databases of YaleB and PIE. SDA aims to
find a projection which respects the \emph{discriminant} structure
inferred from the labeled data points, as well as the intrinsic
\emph{geometric} structure inferred from both labeled and
unlabeled data points. We combine SDA with different graphs to
learn the subspace, and employ the nearest neighbor classifier. We
run the algorithms multiple times with randomly selected data
sets. In each run, 30 images from each subject are randomly
selected as training images, while the rest images as test images.
Among these 30 training images, some images are randomly labeled.
Note here that different from the above transductive
classification, the test set is not available in the subspace
learning stage. Table~\ref{tab:SDA} tabulates the recognition
error rates for different graphs under different labeling
percentages. We can see that NNLRS-graph almost consistently
outperforms other graphs.

\begin{table*}[t]
\caption{Classification error rates (\%) of various graphs
combined with the GHF label propagation method under different
percentages of labeled samples (shown in the parenthesis after the
dataset names). The bold numbers are the lowest error rates under
different sampling percentages.} \label{tab:GHF}
\begin{center}
\begin{tabular}{cccccccccc}
\hline
Dataset & $k$NN0 & $k$NN1 & LLE0 & LLE1 & $\ell_1$-graph & SPG & LRR & NNLRS   & NNLRS-EF\\
\hline
YaleB (10\%) & 33.51 & 38.27 & 29.21 & 29.94 & 46.13  & 15.57 & 28.22 & 3.75  & \textbf{2.37} \\
YaleB (20\%) & 34.66 & 38.97 & 30.63 & 30.63 & 45.54  & 17.56 & 24.46 & 9.84  & \textbf{3.33} \\
YaleB (30\%) & 33.71 & 37.87 & 28.17 & 28.17 & 46.14  & 16.54 & 22.33 & 10.54 & \textbf{2.67} \\
YaleB (40\%) & 33.00 & 37.34 & 28.36 & 28.36 & 43.39  & 17.16 & 19.42 & 9.38  & \textbf{2.44} \\
YaleB (50\%) & 33.10 & 37.38 & 28.38 & 28.38 & 42.25  & 18.99 & 18.04 & 9.64  & \textbf{2.67} \\
YaleB (60\%) & 32.48 & 37.78 & 28.53 & 28.53 & 41.52  & 20.50 & 16.09 & 8.13  & \textbf{2.67} \\
\hline
PIE (10\%) & 34.84 & 37.54 & 33.06 & 33.44 & 22.88 & 20.50 & 33.98 & \textbf{11.11} & 11.83 \\
PIE (20\%) & 37.46 & 40.31 & 35.05 & 35.81 & 22.94 & 20.30 & 34.35 & 22.81 & \textbf{7.12} \\
PIE (30\%) & 35.30 & 37.80 & 32.52 & 32.88 & 22.33 & 20.60 & 31.81 & 17.86 & \textbf{5.6} \\
PIE (40\%) & 35.81 & 38.22 & 32.51 & 32.99 & 23.14 & 20.81 & 32.39 & 16.25 & \textbf{4.5} \\
PIE (50\%) & 34.39 & 37.38 & 31.41 & 31.64 & 23.01 & 21.43 & 31.33 & 19.25 & \textbf{3.8} \\
PIE (60\%) & 35.63 & 38.00 & 32.76 & 32.85 & 25.76 & 23.82 & 32.50 & 21.56 & \textbf{3.79} \\
\hline
\comment{
COIL20 (10\%) & 9.51 & 11.56 & \textbf{9.28} & 9.83 & 29.47 & 10.01 & 11.88 & 13.33     & 1  \\
COIL20 (20\%) & 9.18 & 11.29 & \textbf{8.66} & 9.80 & 29.08 & 9.44  & 10.98 & 13.91     & 1  \\
COIL20 (30\%) & 8.23 & 10.07 & \textbf{7.48} & 8.43 & 27.95 & 9.76  & 9.27  & 9.38      & 1  \\
COIL20 (40\%) & 8.46 & 10.29 & \textbf{7.74} & 8.33 & 27.52 & 10.82 & 8.83  & 9.42      & 1  \\
COIL20 (50\%) & 9.51 & 11.30 & \textbf{8.83} & 9.53 & 30.19 & 14.54 & 10.84 & 11.13     & 1  \\
COIL20 (60\%) & 9.75 & 12.05 & 9.58          & 10.69& 30.81 & 17.30 & 11.45 & \textbf{6.25} & 1  \\
\hline
}

USPS (10\%) & 1.87  & 2.20  & 17.10 & 27.31  & 43.27 & 3.95  & 2.25  & \textbf{1.57} &  2.36 \\
USPS (20\%) & 2.51  & 2.67  & 22.92 & 30.83  & 41.27 & 5.28  & 3.10  & 1.93  &  \textbf{1.90} \\
USPS (30\%) & 5.88  & 6.10  & 21.26 & 27.54  & 38.31 & 10.48 & 8.91  & 4.95  &  \textbf{1.79} \\
USPS (40\%) & 7.87  & 8.44  & 19.21 & 22.78  & 34.86 & 14.22 & 13.44 & 7.44  &  \textbf{1.53} \\
USPS (50\%) & 17.19 & 18.44 & 18.41 & 19.48  & 29.42 & 20.38 & 21.88 & 11.27 &  \textbf{1.58} \\
USPS (60\%) & 11.04 & 15.20 & 14.80 & 14.94  & 23.36 & 15.89 & 17.75 & 6.09  &  \textbf{1.40} \\
\hline
\end{tabular}
\end{center}
\end{table*}

\begin{table*}[tb]
\caption{Classification error rates (\%) of various graphs combined with the LGC label
propagation method under different percentages of labeled samples (shown in the parenthesis
after the dataset names). The bold numbers are the lowest error rates under different
sampling percentages.}
\label{tab:LGC}
\begin{center}
\begin{tabular}{cccccccccc}
\hline
Dataset & $k$NN0 & $k$NN1 & LLE0 & LLE1 & $\ell_1$-graph & SPG & LRR & NNLRS  & NNLRS-EF \\
\hline
YaleB (10\%) & 32.89 & 36.84 & 29.00 & 29.76 & 46.82 & 16.37 & 28.22 & 5.56  & \textbf{2.37}\\
YaleB (20\%) & 31.09 & 35.59 & 25.84 & 26.65 & 50.53 & 12.39 & 24.46 & 5.31  & \textbf{2.67}\\
YaleB (30\%) & 28.56 & 33.54 & 22.24 & 22.83 & 52.33 & 9.57  & 22.33 & 4.29  & \textbf{2.86}\\
YaleB (40\%) & 26.35 & 30.97 & 19.82 & 19.90 & 57.16 & 7.07  & 19.42 & 3.75  & \textbf{2.89}\\
YaleB (50\%) & 24.78 & 29.73 & 17.61 & 17.65 & 65.79 & 5.63  & 18.04 & 4.00  & \textbf{3.2}\\
YaleB (60\%) & 22.98 & 28.58 & 15.75 & 15.94 & 77.56 & 4.42  & 16.09 & \textbf{3.23}  & 3.3\\
\hline
PIE (10\%) & 34.28 & 36.42 & 32.25 & 32.53 & 21.71 & 19.75 & 31.26 & 12.22 & \textbf{11.87}\\
PIE (20\%) & 33.06 & 36.11 & 30.42 & 30.83 & 17.18 & 15.45 & 29.82 & 10.63 & \textbf{7.49}\\
PIE (30\%) & 30.11 & 33.51 & 26.52 & 27.01 & 12.06 & 10.71 & 25.61 & 9.82  & \textbf{5.31}\\
PIE (40\%) & 28.46 & 32.15 & 23.62 & 24.01 & 9.01  & 8.25  & 23.86 & 7.08  & \textbf{4.58}\\
PIE (50\%) & 26.96 & 30.45 & 21.65 & 22.22 & 6.61  & 6.29  & 21.24 & 4.00  & \textbf{3.77}\\
PIE (60\%) & 25.09 & 29.09 & 19.56 & 20.02 & 5.13  & 4.95  & 20.05 & 5.00  & \textbf{3.63}\\
\hline
\comment{
COIL20 (10\%) & 10.36 & 11.71 & \textbf{9.86} & 10.38 & 28.96 & 11.33 & 11.53 & 12.40 & 1\\
COIL20 (20\%) & 7.62  & 9.21  & 6.86 & 7.38  & 23.07 & \textbf{6.38}  & 7.14  & 8.53  & 1\\
COIL20 (30\%) & 6.44  & 7.94  & 4.87 & 5.56  & 17.78 & \textbf{3.43}  & 5.26  & 6.50  & 1\\
COIL20 (40\%) & 6.03  & 7.61  & 4.51 & 5.19  & 15.76 & \textbf{2.27}  & 4.20  & 4.68  & 1\\
COIL20 (50\%) & 6.00  & 8.20  & 4.63 & 5.36  & 14.38 & \textbf{2.03}  & 4.23  & 3.50  & 1\\
COIL20 (60\%) & 6.17  & 8.16  & 4.20 & 5.02  & 11.97 & 1.73  & 3.59  & \textbf{1.25}  & 1\\
\hline
}
USPS (10\%) & 3.13 & 3.21 & 27.69 & 35.06 & 33.52 & 6.92 & 3.49 & 2.80 & \textbf{2.62}\\
USPS (20\%) & 2.22 & 2.10 & 22.43 & 28.96 & 26.42 & 4.04 & 1.83 & 1.62 & \textbf{1.58}\\
USPS (30\%) & 1.55 & 1.53 & 19.18 & 25.30 & 18.92 & 2.69 & 1.22 & 1.13 & \textbf{1.05}\\
USPS (40\%) & 1.20 & 1.18 & 16.62 & 22.53 & 16.64 & 1.88 & 0.92 & 0.88 & \textbf{0.87}\\
USPS (50\%) & 0.82 & 0.86 & 14.28 & 20.01 & 11.67 & 1.14 & 0.61 & 0.59 & \textbf{0.53}\\
USPS (60\%) & 0.65 & 0.72 & 12.61 & 17.69 & 8.89  & 0.83 & 0.49 & \textbf{0.48} & \textbf{0.48}\\
\hline
\end{tabular}
\end{center}
\end{table*}

\begin{table*}[t]
\caption{Recognition error rates (\%) of various graphs for
semi-supervised discriminative analysis under different
percentages of labeled samples.} \label{tab:SDA}
\begin{center}
\begin{tabular}{ccccccccc}
\hline
Dataset & $k$NN0 & $k$NN1 & LLE0 & LLE1 & $\ell_1$-graph & SPG  & LRR & NNLRS\\
\hline
YaleB (10\%) & 43.79 & 48.55 & 39.06 & 39.43 & 37.29 & 36.88 & 40.18 & \textbf{34.46}\\
YaleB (20\%) & 30.31 & 34.37 & 25.30 & 25.59 & 23.87 & 23.56 & 27.96 & \textbf{22.43}\\
YaleB (30\%) & 20.14 & 23.16 & 16.04 & 16.23 & 14.69 & 14.58 & 18.38 & \textbf{14.09}\\
YaleB (40\%) & 13.95 & 16.01 & 10.57 & 10.84 & 9.87 & 9.68 & 12.60 & \textbf{9.40}\\
YaleB (50\%) & 9.89 & 11.69 & 7.34  & 7.42  & 6.78  & 6.78  & 9.03  & \textbf{6.49}\\
YaleB (60\%) & 7.56 & 9.78 & 5.71 & 5.79 & 5.32 & 5.30 & 7.09 & \textbf{5.16}\\
\hline
PIE (10\%) & 44.53 & 48.80 & 38.79 & 39.30 & 35.82 & 35.02 & 42.20 & \textbf{34.40}\\
PIE (20\%) & 29.16 & 33.60 & 23.57 & 24.02 & 21.33 & 20.84 & 27.35 & \textbf{20.74}\\
PIE (30\%) & 16.26 & 19.26 & 12.58 & 12.76 & 11.37 & 11.13 & 15.30 & \textbf{11.11}\\
PIE (40\%) & 10.74 & 13.05 & 8.26  & 8.44  & 7.55  & \textbf{7.42} & 10.28 & 7.47\\
PIE (50\%) & 7.26  & 8.55  & 5.70  & 5.77  & 5.30  & 5.23  & 6.93  & \textbf{5.17}\\
PIE (60\%) & 5.36  & 6.23  & 4.38  & 4.42  & 4.11  & \textbf{4.08} & 5.22  & \textbf{4.08}\\
\hline
\comment{
COIL20 (10\%) & 25.30 & 27.03 & 24.73 & 25.09 & 26.96 & 25.43 & 49.77 & \textbf{23.62}\\
COIL20 (20\%) & 15.73 & 17.20 & 14.97 & 15.24 & 15.90 & 15.08 & 34.70 & \textbf{13.77}\\
COIL20 (30\%) & 9.50  & 11.23 & 8.31  & 8.45  & 8.76  & 8.36  & 22.09 & \textbf{7.12}\\
COIL20 (40\%) & 6.19  & 7.63  & 5.38  & 5.50  & 5.60  & 5.33  & 14.98 & \textbf{4.18}\\
COIL20 (50\%) & 4.30  & 5.43  & 3.66  & 3.76  & 3.69  & 3.65  & 10.05 & \textbf{2.54}\\
COIL20 (60\%) & 3.03  & 3.77  & 2.61  & 2.64  & 2.70  & 2.63  & 7.00  & \textbf{1.67}\\
\hline
}
\end{tabular}
\end{center}
\end{table*}

\subsection{Parameters Sensitivity of NNLRS-graph}
In this subsection, we examine the parameter sensitivity of
NNLRS-graph, which includes two main parameters, $\beta$ and
$\lambda$. $\beta$ is to balance the sparsity and the
low-rankness, while $\lambda$ is to deal with the gross corruption
errors in data. Large $\beta$ means that we emphasize the sparsity
property more than the low-rankness property. We vary the
parameters and evaluate the classification performance of
NNLRS-graph based SDA on the PIE face database. Since the
percentage of gross corruption errors in data should be fixed, we
set $\lambda = 10$ empirically\footnote{In the above experiments,
we did not tune $\lambda$ either.} and only vary $\beta$. Because
here we test many parametric settings, like the above experiments
here we only average the rates over 5 random trials. The results
are shown in Table~\ref{tab:W}. From this table, we can see that
the performance of NNLRS-graph based SDA decreases when $\beta >
1$. If we ignore the sparsity property (i.e., $\beta = 0$), the
performance also decreases. This means that both sparsity property
and low-rankness property are important for graph construction. An
informative graph should reveal the global structure of the whole
data and be as sparse as possible. In all of our experiments
above, we always set $\beta = 0.2$.
\begin{table*}[tb]
\caption{Recognition error rates (\%) of NNLRS-graph for
semi-supervised discriminative analysis on the PIE face database
under different percentages of labeled samples. $\lambda$ is fixed
at 10.} \label{tab:W}
\begin{center}
\begin{tabular}{cccccccccc}
\hline
$\beta$ & 0     & 0.001 & 0.01  & 0.2   &  0.8  & 1     & 5     & 10    &  100\\
\hline
  10\%  & 38.93 & 26.48 & 26.48 & 26.43 & 26.62 & 27.05 & 39.10 & 39.52 & 40.24\\
  20\%  & 24.03 & 20.10 & 20.10 & 20.05 & 20.05 & 20.24 & 28.90 & 30.76 & 31.81\\
  30\%  & 13.56 & 12.10 & 12.10 & 12.19 & 12.19 & 12.19 & 16.57 & 16.95 & 18.48\\
  40\%  & 9.48  & 8.14  & 8.14  & 8.14  & 8.10  & 8.14  & 13.05 & 13.38 & 13.33\\
  50\%  & 6.57  & 5.48  & 5.52  & 5.52  & 5.57  & 5.57  & 7.19  & 6.71  & 6.52\\
  60\%  & 6.10  & 5.86  & 5.86  & 5.86  & 5.86  & 5.86  & 6.76  & 7.10  & 7.95\\
\hline
\end{tabular}
\end{center}
\end{table*}

\begin{figure}[tb]
\centering
\subfloat[Classification error rate on the YaleB face database]{
\label{fig:YaleB}
\begin{minipage}[t]{0.45\textwidth}
   \centering
   \includegraphics[width = 1\textwidth]{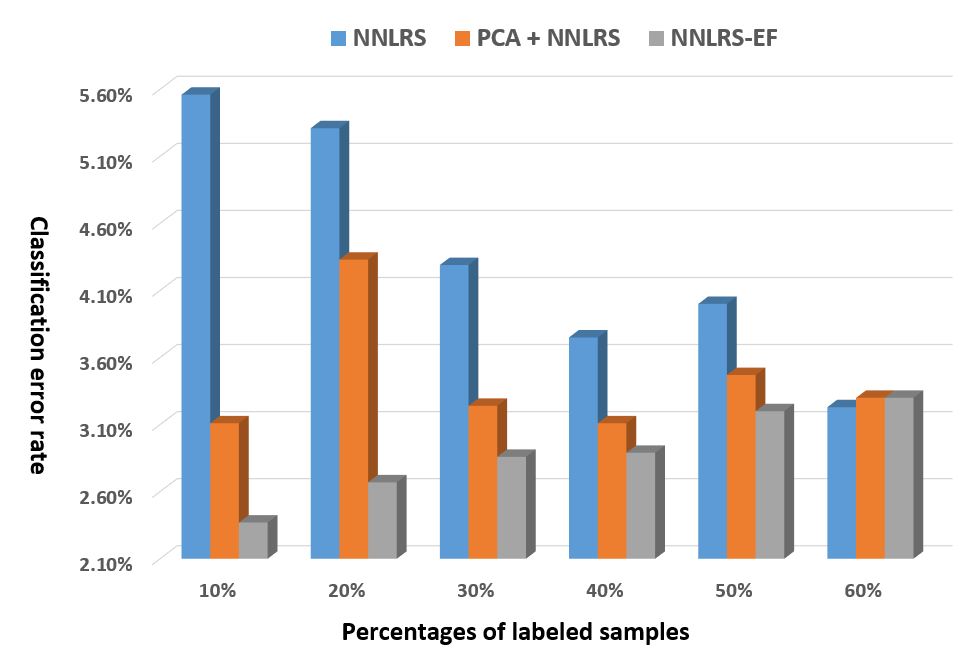}
\end{minipage}
} \\
\subfloat[Classification error rate on the PIE face database]{
\label{fig:PIE}
\begin{minipage}[t]{0.45\textwidth}
   \centering
   \includegraphics[width = 1\textwidth]{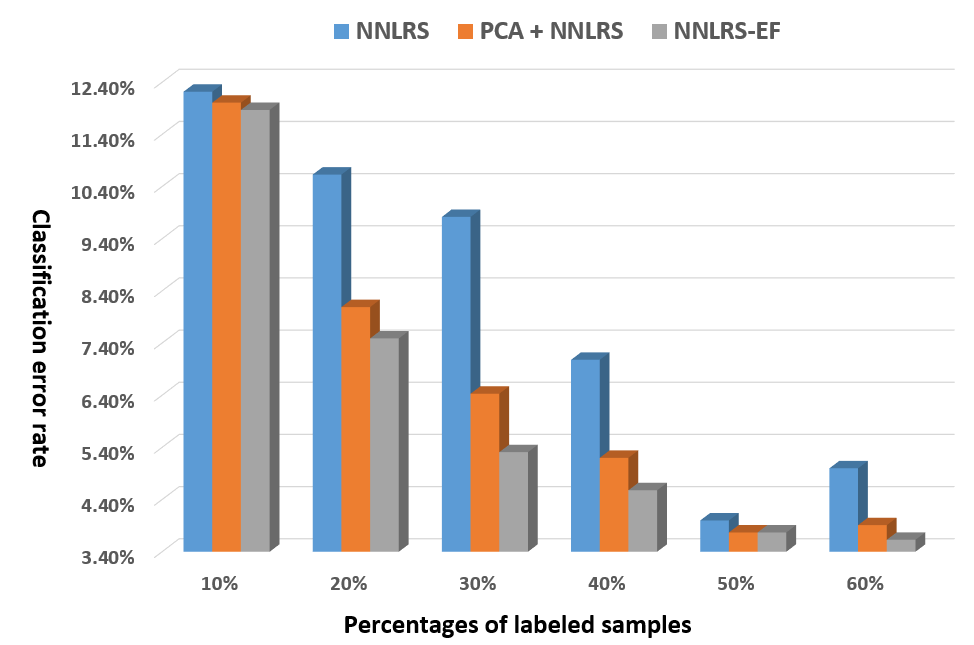}
\end{minipage}
}
\caption{Classification error rate on the YaleB face database and
the PIE face database, using the LGC label propagation method
under different percentages of labeled samples.}
\label{fig:jointly_learning}
\end{figure}

\subsection{Joint Learning vs. Independent Learning}
In this subsection, we further examine the effectiveness of joint
feature learning. As our feature learning method is mostly related
to PCA, we propose to compare our method with the following
baseline. We first reduce the dimensionality of the data with PCA.
Then we use the embedded data by applying PCA. For fair
comparison, we also keep the dimensionality of the data to be 100,
which is exactly the same as that of our NNLRS-EF. For simplicity,
we denote such baseline method as PCA+NNLRS. We show the
performance of NNLRS, PCA+NNLRS, and NNLRS-EF in semi-supervised
learning in Fig.~\ref{fig:jointly_learning}. In all the
experiments, we keep the same setting. From this figure, we can
have the following observations:
\begin{itemize}
\item The performance of NNLSR with embedded data as features (PCA+NNLRS and NNLRS-EF)
is better than that of NNLRS with raw pixels. This observation demonstrates
the necessity of data embedding for data structure discovery.
\item The performance of NNLRS-EF is better than PCA+NNLRS which
does PCA and learns the NNLRS separately. Such an observation
proves that joint learning can learn more proper data
representation for the subsequent data structure discovery, which
demonstrates the effectiveness of our NNLRS-EF framework.
\end{itemize}

\section{Conclusion}
\label{sec:conclusion} This paper proposes a novel informative graph,
called the nonnegative low rank and sparse graph
(NNLRS-graph), for graph-based semi-supervised learning.
NNLRS-graph mainly uses two important properties of
high-dimensional data, sparsity and low-rankness, both of which
capture the structure of the whole data. It simultaneously derives
the graph structure and the graph weights, by solving a problem of
nonnegative low rank and sparse representation of the \emph{whole}
data. Extensive experiments on both classification and
dimensionality reduction show that, NNLRS-graph is better at
capturing the globally linear structure of data, and thus is more
informative and more suitable than other graphs for graph-based
semi-supervised learning. Moreover, as good features are robust to
data noise and thus help to reveal the relationship among data
points, we propose to include the data embedding and construct the
graph within one framework. Experiments show that joint feature
learning does significantly improve the performance of
NNLRS-graph.



\bibliographystyle{IEEEtran}
\bibliography{egbib}

\end{document}